\documentclass[runningheads,a4paper]{llncs}

\usepackage{times}
\usepackage{amssymb}
\usepackage{amsmath}
\usepackage{graphicx}
\usepackage{url}
\usepackage{xcolor}
\usepackage[colorinlistoftodos]{todonotes}

\newcommand{\keywords}[1]{\par\addvspace\baselineskip
\noindent\keywordname\enspace\ignorespaces#1}

\urldef{\mailsa}\path|{ melika.golestani , borhanifardz , hfaili } @ut.ac.ir|
\urldef{\mailsb}\path|srazavi@cs.rochester.edu|
\urldef{\mailsd}\path|ftahmas@emory.edu|

\begin{document}

\title{Using BERT Encoding and Sentence-Level Language Model for Sentence Ordering}

\titlerunning{Using BERT Encoding and Sentence-Level Language Model for Sentence Ordering}

\author{Melika Golestani \inst{1} \and Seyedeh Zahra Razavi\inst{2} \and Zeinab Borhanifard\inst{1} \and Farnaz Tahmasebian \inst{3} \and Hesham Faili\inst{1}}


\authorrunning{Melika Golestani et al.}

\institute{School of Electrical and Computer Engineering, College of Engineering, University of Tehran, Tehran, Iran \\ 
\mailsa\\
Department of Computer Science, University of Rochester, Rochester, USA \\
\mailsb\\
Department of Computer science, Emory university, Atlanta, USA \\
\mailsd\\
}

\index{Golestani, Melika}
\index{Razavi, Seyedeh Zahra}
\index{Borhanifard, Zeinab}
\index{Tahmasebian, Farnaz}
\index{Faili, Hesham}

\toctitle{} \tocauthor{}

\maketitle

%
%
%
%
\begin{abstract}
Discovering the logical sequence of events is one of the cornerstones in Natural Language Understanding. One approach to learn the sequence of events is to study the order of sentences in a coherent text. Sentence ordering can be applied in various tasks such as retrieval-based Question Answering, document summarization, storytelling, text generation, and dialogue systems. Furthermore, we can learn to model text coherence by learning how to order a set of shuffled sentences.
Previous research has relied on RNN, LSTM, and BiLSTM architecture for learning text language models. However, these networks have performed poorly due to the lack of attention mechanisms. We propose an algorithm for sentence ordering in a corpus of short stories. Our proposed method uses a language model based on Universal Transformers (UT) that captures sentences' dependencies by employing an attention mechanism. Our method improves the previous state-of-the-art in terms of Perfect Match Ratio (PMR) score in the ROCStories dataset, a corpus of nearly 100K short human-made stories. The proposed model includes three components: Sentence Encoder, Language Model, and Sentence Arrangement with Brute Force Search. 
The first component generates sentence embeddings using SBERT-WK pre-trained model fine-tuned on the ROCStories data. Then a Universal Transformer network generates a sentence-level language model. For decoding, the network generates a candidate sentence as the following sentence of the current sentence. We use cosine similarity as a scoring function to assign scores to the candidate embedding and the embeddings of other sentences in the shuffled set. Then a Brute Force Search is employed to maximize the sum of similarities between pairs of consecutive sentences.

\end{abstract}

\section{Introduction}
Modeling text coherence and logical sequences of events is a fundamental problem in natural language processing \cite{logeswaran2018sentence}. Coherence represents the logical connections between the words, sentences and events in a text. An event is generally considered to be the verb of a sentence along with its constellation of arguments, such as subject and object \cite{chambers2009unsupervised}. In a coherent narrative text, events follow a logical order, which makes it possible for a reader to make sense of the text. In NLP tasks, ``Sentence Ordering'' models are suggested to learn high-level structure that causes sentences to appear in a specific order in human-authored texts and by learning to order sentences we can model text coherence\cite{logeswaran2018sentence}.

``Sentence ordering'' refers to organizing a given shuffled set of sentences into a coherent order \cite{logeswaran2018sentence}. Sentence ordering assists in modeling coherence in text, which in turn can be applied in a variety of tasks such as multi-document summarization \cite{barzilay2002inferring,lapata2003probabilistic}, story understanding \cite{mostafazadeh2016caters}, and retrieval based QA systems \cite{logeswaran2018sentence}. A key factor to to solve the problem of sentence ordering is to find a coherent sequence of events appeared in the input sentences. In natural language processing, events are often meant to be the verbs of sentences and their dependencies, such as the subject and the object \cite{mostafazadeh2017event}. A number of studies have proposed methods to order events based on their temporal and causal relations \cite{mostafazadeh2016caters}, or the semantic roles \cite{chambers2008unsupervised}. The researchers developed a Skip-thought \cite{kiros2015skip} encoder-decoder model utilizing an RNN. It takes a sentence and encodes it into a constant-length vector, followed by previous and subsequent sentences. Therefore, all input is first encrypted using a hidden network vector; then its output is decrypted from that vector. Pichotta \& Mooney \cite{pichotta-mooney-2016-using} replaced the RNN with an LSTM, and proposed a sentence-level language model to predict the next sentence for  the script inference task. Nevertheless, skip-thought's architecture's simplicity results in a lower quality of creating sentences embeddings compared to many newer methods and networks like BERT \cite{devlin-etal-2019-bert}. The system uses an RNN-based encoder-decoder architecture, whereas SBERT-WK \cite{Wang2020SBERTWKAS} uses a Transformer-based architecture and an attention mechanism to encode the sentences into vectors.

According to \cite{chambers2008unsupervised}, a statistical script system model an event sequence as a probabilistic model and infer additional events from a document events. This system teaches the linguistic model of events; despite this, some critical information in the sentence can be ignored because only verbs and their dependents are used. Thus, the clue words (adverbs such as ``then'', ``before'', and ``after'') are omitted.

In this paper, we propose a sentence-level language model to solve the problem of sentence ordering. Language models have been popular for years at the word and letter levels \cite{woodland2002large,kozielski2013open}. LSTMs \cite{hochreiter1997long} and RNNs \cite{medsker2001recurrent} have been vastly used for teaching language models at the word and letter levels is beneficial \cite{mikolov2011empirical,jozefowicz2016exploring,jozefowicz2016exploring,10.5555/2969033.2969173,bengio2003neural}. These models take a single word or character as input at time t, update the hidden mode vector, and predict the next word or character at time t+1. Kiros used a sentence level language model based on the RNN \cite{kiros2015skip}. Pichotta in \cite{pichotta-mooney-2016-using} used LSTM-based sentence-level models for script inference due to the problem of RNNs, the vanishing gradients. RNNs and LSTMs encrypt information about one sentence and decode information about the following sentence. Despite this, these networks not use a self-attention mechanism\cite{parikh-etal-2016-decomposable} for encoding sentences, and some information is lost.

We use a sentence level language model based on the universal transformers (UT) \cite{dehghani2018universal}. A generalization of the Transformer model, the UT model is a parallel-in-time recurrent sequence model. The UT model combines two aspects of feed-forward models: parallelism and global receptive field.  An additional feature is the position-based halting mechanism. Using a self-attention mechanism, the Universal Transformer refines its representations for all sentences in the sequence in parallel.

Our model involves three components: the first component encodes sentences; the second one uses a neural network to teach the language model at the sentence level by universal transformers (UT) and to predict the next sentence. The last component uses the cosine similarity and Brute Force search to order sentences. We aim to design a model that can capture the sequence of events by arranging a set of sentences.

The model we propose is designed to organize shuffled sentences of short stories. By using this method, we encode sentences by the SBERT-WK model, which causes us to pay attention to every word within a sentence equal to its importance and transform the sentence into a vector. Thus the embedding contains the clue words information. To capture the relationship between Sentences, We use a UT-based language model, which predicts the next sentence of the current sentence. Finally, we use a Brute Force search, using the scoring function based on cosine similarity, which is scored between the candidate sentence and other sentences, to maximize the sum of similarities between two consecutive sentences. As dataset, we use ROCStories \cite{mostafazadeh2016corpus}, which includes around 100K short 5-sentence stories, written by human turkers. The gold data follows the order similar to the original order in the stories. 

The main applications of sentence ordering can be mentioned as: extractive and multi-document text summarization \cite{barzilay2002inferring}, retrieval based QA systems \cite{logeswaran2018sentence}, storytelling \cite{mostafazadeh2016corpus}, text coherence modeling \cite{logeswaran2018sentence}, discourse coherence \cite{elsner2007unified}, and text generation systems \cite{bosselut-etal-2018-discourse}.


\section{Related Work}
\label{Related Work}
Understanding relations between sentences has become increasingly important for various NLP tasks, such as multi-document summarization \cite{barzilay2002inferring}, text generation \cite{bosselut-etal-2018-discourse}, and text coherence modeling \cite{logeswaran2018sentence}. Moreover, learning the order of sentences can help in modeling text coherence \cite{logeswaran2018sentence}. In sentence ordering, the goal is to arrange a set of unordered sentences in a cohesive order. Ordering models aim to identify patterns resulting sentences to appear in a specific order in a coherent text. Several previous studies have addressed the task of sentence ordering for a set of data such as news articles \cite{barzilay2002inferring,lapata2003probabilistic,bollegala2010bottom}. 

Previous methods suggested on the task of sentence ordering fall into two categories: traditional approaches, and deep learning-based approaches. Traditional approaches to coherence modeling and sentence ordering often apply probabilistic models \cite{lapata2003probabilistic}. Barzilay utilized content models to represent topics as HMM states \cite{barzilay-lee-2004-catching}, then employed hand-crafted linguistic features, including Entity Grid, to model the document structure \cite{barzilay2008modeling}. \cite{mostafazadeh2016corpus} suggested using n-gram overlapping to pick a final sentence from two human-made options of the ROCStories stories. In \cite{pour2020new}, a similar method is used to arrange the set of unordered sentences of the stories. As the next sentence, they pick the sentence with the most n-gram overlaps with the current sentence. They measured an overlap up to 4-gram using Smoothed-BLEU \cite{lin2004automatic}.

In recent years, researchers used neural approaches to solve sentence ordering tasks. \cite{agrawal-etal-2016-sort} proposed SkipThought + Pairwise model. The method involves combining two points identified from the unary embedding of sentences without considering context and a pairwise model of sentences based on their relative composition in context. However, as SkipThought model \cite{kiros2015skip} was applied to map sentences into vector space, some information is missing. Another method proposed in \cite{li-jurafsky-2017-neural} is a productive model that is called Seq2Seq + Pairwise. A Seq2Seq model based on an encoder-decoder architecture is used to predict the following sentence, having one. This way, the model learns a probabilities distributions over sentences.

Gong employed an end-to-end sentence ordering method, LSTM + PtrNet\cite{gong2016end}. After getting encrypted and decoded by LSTM models, pointer networks are being used to arrange sentences. Another method proposed in \cite{logeswaran2018sentence}, called LSTM + Set2Seq, uses LSTM and attention mechanism to encode sentences and learn a representation of context. Then a pointer network is applied to ranks sentences. 
\cite{wang2019hierarchical} proposed Hierarchical attention networks (HAN). The idea is to capture the information of clue words to learn the dependency between sentences. The system uses a word encoder with a BiLSTM architecture, an attention layer that creates sentence embedding, and an attention mechanism to encode and decode a group of sentences together. Although LSTM and BiLSTM-based models show improvement, they are less efficient comparing to a BERT-based encoder model. This is mostly because of the fact that BiLSTM models do not look at both directions of a word simultaneously. In contrast, BERT is trained to learn both left and right positions at the same time.


\cite{journals/corr/abs-1301-3781} proposed a CBoW model that generate sentence embedding using the average of the word embedding vectors of the words that compose the sentence.
This embedding is used in \cite{pour2020new} to suggest a baseline, where cosine similarity between embeddings of pairs of sentences is used to order the pairs. This has two essential defects: it does not include many sentence information such as clue words and there is no way for modelling sentences' dependencies. 

\cite{pour2020new} presents sentence correlation measure (SCM) for sentence ordering. This measure has three main components:
\begin{enumerate}
\item A sentence encoding component based on the pretrained SBERT-WK \cite{Wang2020SBERTWKAS};
\item A scoring component based on the cosine similarity;
\item  A ranking component based on a Brute Force search. 
\end{enumerate}
None of the above components need access to training data which provides an advantage where limited data is available\cite{pour2020new}. The method does not work as well as HAN or LSTM + Set2Seq when enough training data is available, since dependencies between sentences are not captured.

\cite{kumar2020deep} proposed RankTxNet ListMLE, which is a  pointer-based model. This method relies on a pre-trained BERT model for encoding sentences and a self-attention based transformer network for encoding paragraphs. To predict a relative score for each sentence, they use a feed-forward network as a decoder which determines each sentence's position in a paragraph.

Our proposed model has some similarity with the methods presented in \cite{gong2016end} and \cite{logeswaran2018sentence} in using neural networks for for predicting sentence ordering in a pairwise manner. However, we employ a BERT-based model for sentence encoding, in contrast to the LSTM-based model they used. Moreover, by applying universal transformers we encode a group of sentences and learn a language model.

\section{Methodology}
\label{Methodology}
\subsection{Task Formulation}
\label{Task Formulation}
The sentence ordering takes the story S as input, which its sentences are probably unsorted:
\begin{equation}
S : {s_1, s_2, ..., s_n} 
\label{eq:1}
\end{equation}
Where n is the length of the sequence or the number of sentences in each sequence. It outputs a permutation of the sentences like o  so that o is equal to \begin{math} o^* \end{math}: 
\begin{equation}
s(o_1^*) > s(o_2^*) > \ldots > s(o_5^*),
\label{eq:2}
\end{equation}
where \begin{math} o^* \end{math} is to the order of the sentences in the gold data.

\subsection{Sentence-Level Language Model for Sentence Ordering (SLM)}
\label{SLM}
\begin{figure}
\includegraphics[width=\textwidth]{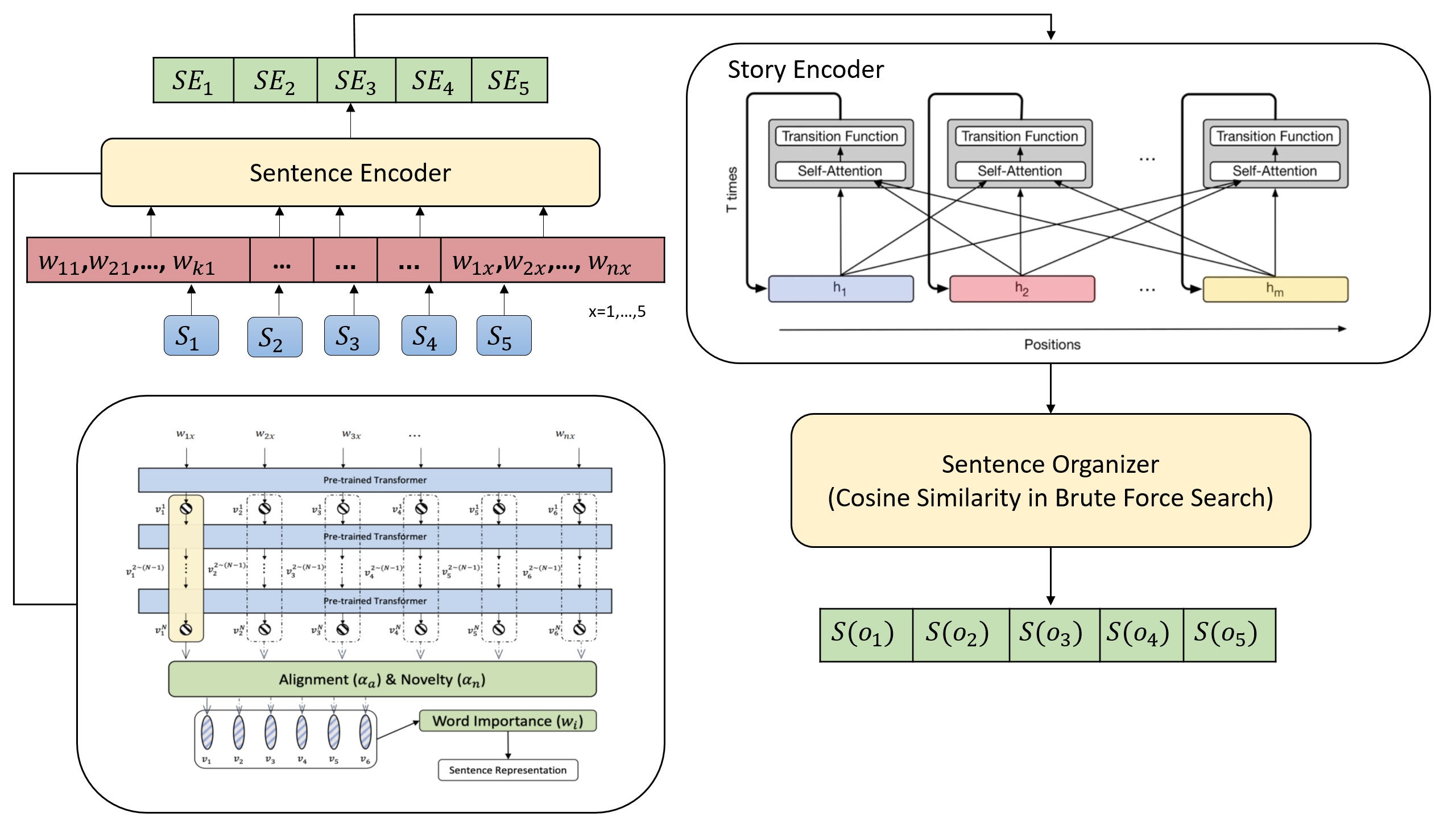}
\caption{Showing the SLM in the abstract. During model training, sentence ordering is given to the network correctly to learn and model the relationships and dependencies between the sentences. During testing, input sentences are shuffled. Where $w_{j_i}$ is the jth word in the ith sentence and $s(o_i)$ is the ith sentence in output order. The goal is to output an s(o) equal to $s(o^*)$ (this represents the gold data ordering).}
\label{figure 1}
\end{figure}

We proposed Sentence-level Language Model (SLM) for Sentence Ordering compose of Sentence Encoder, Story Encoder, and Sentence Organizer. The architecture of the proposed model is shown in Figure\ref{figure 1}.
Sentence encoder gets sentence as input and encoded it into a vector using a fine-tuned pre-trained SBERT-WK\cite{Wang2020SBERTWKAS}. The embedding pays more attention to the sentence's crucial parts, such as the verbs and clue words. Furthermore, stop words are omitted. Then, Story Encoder takes the sentence encoder as an input and learns the sentence level language model using an encoder-decoder architecture based on universal transformers (UT) \cite{dehghani2018universal}. UT couples parallelism with the global receptive field in a feed-forward model. The UT model also includes position-based halting \cite{dehghani2018universal}. The UT-based encoder-decoder component corrects vector representations for each position (sentence) in parallel by combining information from different positions using self-attention and applying a repetitive transfer function \cite{dehghani2018universal}. Hence dependencies between the sentences are captured. The vector is learned from the hidden state and then decodes the vector and indicates the next sentence's candidate. 

What matters at runtime is that during training, information are transferred to all nodes at time $t+1$ from all nodes at time $t$. Thus, parallelism is created, which allows all tokens processed at the same time \cite{dehghani2018universal}.

In the Sentence Organizer, we can rate each sentence and the candidate based on their similarity. We use the cosine similarity as the scoring function. The arrangement is made using a Brute Force search in all state spaces to maximize the sum of the cosine similarity between two consecutive sentences like the sentence ordering component in \cite{pour2020new}, as follow:
\begin{equation}
\sum_{i=1, \ldots, 4}S\left(s(o_i), s(o_{i + 1}\right) = \max\sum_{\substack{i, j = 1, \ldots, 5 \\ i \neq j}}S(s'_i,s_j),
\end{equation}
where $S(s'_i, s_j)$ represents the cosine similarity between $s'_i$ and $s_j$, and $s'_i$ is the candidate for the next sentence of the $i$-th sentence, represented with $s_i$. $s(o_i)$ represents the $i$-th sentence in the order of output, and $s_i$ is the $i$-th sentence in the non-ordered permutation of sentences or the input's order.

Our model is trained again using the universal sentence encoder (USE) embeddings \cite{cer-etal-2018-universal} to compare them to SBERT-WK embeddings, and additionally, we employ the BiLSTM \cite{schuster1997bidirectional} network to learn the LM. Also, we apply the nearest neighbor search, which is a greedy algorithm. The NN search calculations for sentence ordering are presented in \cite{pour2020new}. Consequently, we train, evaluate, and test nine different models, including the following:
\begin{itemize}
\item Fine-tuned pretrained SBERT-WK + UT + BFS,
\item Pretrained SBERT-WK + UT + BFS,
\item Pretrained SBERT-WK + BiLSTM + BFS,
\item Pretrained SBERT-WK + UT + NN,
\item Pretrained SBERT-WK + BiLSTM + NN,
\item USE + UT + BFS,
\item USE + BiLSTM + BFS,
\item USE + UT + NN,
\item USE + BiLSTM + NN,
\end{itemize}

So we can assess the effect of change on each component.

\section{Experiment}
\label{Experiment}
\subsection{Dataset}
\label{Dataset}
We used ROCStories dataset (a commonsense story
dataset) \cite{mostafazadeh2016corpus}. The dataset contains 98,162 five-sentence stories with an average word count of 50 words. 3,742 of stories have two choices as the fifth sentence or the final sentence. The corpus has been presented for a shared task called LSDSem \cite{mostafazadeh2017lsdsem} where models are supposed to choose the correct ending to a four-sentence story. All the stories and options are generated by human. This dataset has some essential characteristics that make it a fit for our task of learning sequences of events: ROCStories contains causal and temporal relationships among daily events. This makes it possible to learn the narrative structure of a wide range of events. The dataset contains a collection of daily non-fictional short stories suitable for the training of coherent text models.

\subsection{Baselines and competitors}
\label{Baselines and competitors}

The proposed method is compared with five baselines as follows: Sentence n-gram overlap \cite{pour2020new}, SkipThought + Pairwise \cite{agrawal-etal-2016-sort}, Seq2Seq + Pairwise \cite{li-jurafsky-2017-neural}, Continues Bag of Words (CBoW) \cite{pour2020new}, Sentence Correlation Measure (SCM) \cite{pour2020new}. The competitors methods follow as: Hierarchical Attention Networks (HAN) \cite{wang2019hierarchical}, LSTM+PtrNet \cite{gong2016end}, LSTM+Set2Seq \cite{logeswaran2018sentence}, and RankTxNet ListMLE \cite{kumar2020deep}.

\subsection{Metrics}
\label{Metrics}

The metrics we use to evaluate story ordering outputs are ``Kendal's Tau'' and ``PMR'', introduced below.
\begin{enumerate}
\item Kendall’s Tau ($tau$)

Kendal's Tau measures the quality of arrangements by Equation \eqref{4}, 
\begin{equation}
\tau = 1-((2 \cdot \mathit{number\: of\: inversions}))/(N\cdot(N-1)/2)
\label{4}
\end{equation}
where \(N\) represents the sequence length, and \(number\: of\: inversions\) is equal to how many binaries' relative order is wrongly predicted. This measure for sentence ordering is correlated with human judgment, according to \cite{lapata2006automatic}.

The value of this criterion is in the range \([-1,1]\); the lower limit indicates the worst case, and \(\tau = 1\) when the predicted order equals to the order in the gold data are the same.

\item Perfect Match Ratio (PMR)

PMR defined as following, calculates the ratio of exactly matching sentence orders without penalizing incorrect ones.  
\begin{equation}
\mathrm{PMR}=((\mathit{\#\:of\:Correct\: Pairs}))/(N\cdot(N-1)/2)
\label{5}
\end{equation}

where \(N\) represents the sequence length. One can see that the PMR is always in the range of 0 and 1, where PMR = 1 indicates the predicted order is exactly the same as the gold order, and a PMR = 0 means that the predicted order is precisely the contrary of the gold order.

\end{enumerate}

\subsection{Results and Analysis}
\label{Results and Analysis}

As mentioned in section \ref{Methodology}, SLM has three components. First, a sentence encoder, which is designed based on SBERT-WK. Second, a story encoder, which trains a sentence-level language model and learns the dependencies between the sentences. In this component, the UT network is used. The third component is sentences' organizer which calculates the cosine similarity between sentence embedding vectors and employs a Brute Force search to maximize the sum of similarities among consecutive sentences.

In addition to SLM, we trained and tested nine other models by replacing each component with other algorithms or architectures to select the best algorithm and compare SLM with them. These nine models were constructed by replacing  SBERT-WK encoding with USE vectors, the UT network with a BiLSTM one, and the BFS with a nearest neighbor search. 

The parameters of the SLM are given in Table \ref{1}. The BERT vectors have 768 dimensions, while the USE ones have 512 dimensions. Following previous works \cite{logeswaran2018sentence,wang2019hierarchical}, we randomly split the dataset into training (80\%), test (10\%), and validation (10\%) sets, where they contain 392645, 49080, and 49080 sentences respectively.

\begin{table}[htbp]
\centering
\caption{Parameters of the SLM. SLM encodes sentences with SBERT-WK embeddings and learns a language model with the UT network. Also, the USE embeddings with d$=$512 instead of the SBERT-WK ones are used to allow for a better choice. So hidden layer size of that is 4$\times$ 512.}
\begin{center}
\begin{tabular}{|c|c|}
\hline
\textbf{Method Components }&\textbf{Tau}\\
\hline
\textbf{Initial Learning Rate ($\alpha$)} & \textit{0.5}\\
\hline
\textbf{Regularization ($\lambda$)} & \textit{\begin{math}10^{-5}\end{math}}\\
\hline
\textbf{size of Embedding vectors (d)} & \textit{768}\\
\hline
\textbf{Hidden Layer size (h)} & \textit{4$\times$768}\\
\hline
\end{tabular}
\label{1}
\end{center}
\end{table}

Based on table \ref{2}, SBERT-WK outperforms USE due to its better architecture for sentence encoding. In all cases, UT is superior to BiLSTM, which was expected as UT has an attention mechanism. Due to NN's greedy nature, NN may not always find the global optimal solution and may become stuck in the local optimal, whereas BFS never will. So BFS is better than NN. That is why the best results are happening where we use fine-tuned SBERT-WK embedding, UT to capture dependencies between sentences, cosine similarity as a scoring function, and searching the entire state space to maximize the total similarity of consecutive sentences.

\begin{table}[htbp]
\centering
\caption{SLM results by changing each component.}
\begin{center}
\begin{tabular}{|c|c|c|}
\hline
\textbf{Method Components }&\textbf{Tau}&\textbf{PMR} \\
\hline
\textbf{SLM (Fine tuned SBERT-WK + UT + Brute Force Search)} & \textit{0.7547}& \textit{0.4064} \\
\hline
\textbf{SBERT-WK + UT + Brute Force Search} & \textit{0.7465}& \textit{0.3893} \\
\hline
\textbf{USE + UT + Brute Force Search} & \textit{0.7206}& \textit{0.373}  \\
\hline
\textbf{SBERT-WK + BiLSTM + Brute Force Search} & \textit{0.7317}& \textit{0.3762}  \\
\hline
\textbf{USE + BiLSTM + Brute Force Search} & \textit{0.7044}& \textit{0.3545}  \\
\hline
\textbf{SBERT-WK + UT + Nearest Neighbor Search} & \textit{0.64}& \textit{0.2755}  \\
\hline
\textbf{USE + UT + Nearest Neighbor Search} & \textit{0.6162}& \textit{0.267}  \\
\hline
\textbf{SBERT-WK + BiLSTM + Nearest Neighbor Search} & \textit{0.6214}& \textit{0.2576}  \\
\hline
\textbf{USE + BiLSTM + Nearest Neighbor Search} & \textit{0.5980}& \textit{0.2511}  \\
\hline
\end{tabular}
\label{2}
\end{center}
\end{table}

\begin{figure}
    \centering
    \includegraphics[width=0.8\textwidth]{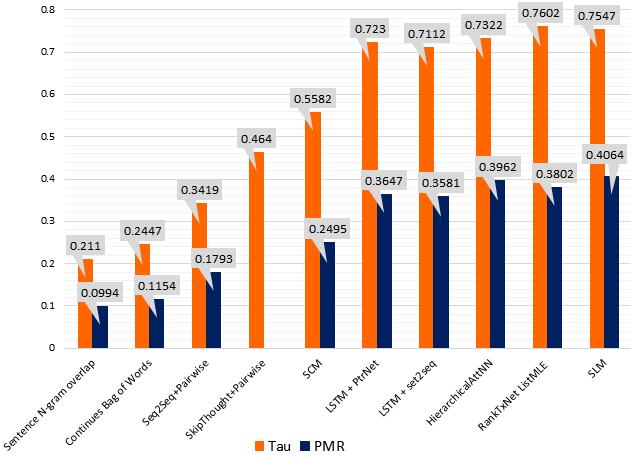}

    \caption{Kendall’s tau ($\tau$) and perfect match ratio (PMR) on test set for ROCStories datasets. The results of the model, compared to the baselines and competitors.}
    \label{figure 2}
\end{figure}

Figure \ref{figure 2} shows the results of the proposed model, SLM, along with the baselines and competitors. SLM has improved by about 3.2\% and 4.2\% compared to LSTM + PtrNet, and more than 4.3\% and 4.8\% compared to LSTM + Set2seq in $\tau$ and PMR, respectively. This performance improvement, is due to the use of a SBERT-WK model sentence encoders and employing an attention mechanism at the story encoder utilizing universal transformers. Taking advantage of the mentioned components, SLM can capture the ``intradependecy'' and ``interdependency'' of sentences very well. Intradependency of a sentence refers to the relations among each sentence's words, while interdependency refers to each sentence's relations with other sentences of the story. Our method is also superior to the HAN network, increasing $\tau$ criterion by more than 2.2 percent and PMR criterion by 1 percent. This could be because of the fact that the sentences are encoded using the BERT-based model in the SLM. HAN, however, uses a BiLSM encoder and multi-head attention layer. 

Besides, SLM performed about 2.6 percent better than RankTxNet in PMR. However, the Kendall's $\tau$ score gives  RankTxNet a slight superiority. The difference is not significant, and can happen as a result of random parameters of the network \cite{sennrich-etal-2016-neural,haddad2018handling}.

\section{Conclusion}
We presented a language-model-based framework to solve sentence ordering task. A sequence of shuffled sentences is inputted to our framework, where the output should be a coherent order of the given set. We achieved state-of-the-art performance in PMR scores on ROCStories dataset. We learned that SBERT-WK is a suitable choice for sentence encoding. We analyzed how the method is changed by using the USE to encode sentences. We also found that Universal Transformers perform better for story encoding and learning the sentence-level language model comparing to BiLSTM models. Moreover, we compared Brute Force Search with Nearest Neighbor Search to order sentences. As future work, we plan to develop more robust models to fulfil ordering of instances with longer input sequences.

\bibliographystyle{splncs04}
\bibliography{lnai12848-paper27}

\begin{thebibliography}{10}
\providecommand{\url}[1]{\texttt{#1}}
\providecommand{\urlprefix}{URL }
\providecommand{\doi}[1]{https://doi.org/#1}

\bibitem{agrawal-etal-2016-sort}
Agrawal, H., Chandrasekaran, A., Batra, D., Parikh, D., Bansal, M.: Sort story:
  Sorting jumbled images and captions into stories. In: Proceedings of the 2016
  Conference on Empirical Methods in Natural Language Processing. pp. 925--931.
  Association for Computational Linguistics, Austin, Texas (Nov 2016).
  \doi{10.18653/v1/D16-1091}, \url{https://www.aclweb.org/anthology/D16-1091}

\bibitem{barzilay2002inferring}
Barzilay, R., Elhadad, N.: Inferring strategies for sentence ordering in
  multidocument news summarization. Journal of Artificial Intelligence Research
   \textbf{17},  35--55 (2002)

\bibitem{barzilay2008modeling}
Barzilay, R., Lapata, M.: Modeling local coherence: An entity-based approach.
  Computational Linguistics  \textbf{34}(1),  1--34 (2008)

\bibitem{barzilay-lee-2004-catching}
Barzilay, R., Lee, L.: Catching the drift: Probabilistic content models, with
  applications to generation and summarization. In: Proceedings of the Human
  Language Technology Conference of the North {A}merican Chapter of the
  Association for Computational Linguistics: {HLT}-{NAACL} 2004. pp. 113--120.
  Association for Computational Linguistics, Boston, Massachusetts, USA (May 2
  - May 7 2004), \url{https://www.aclweb.org/anthology/N04-1015}

\bibitem{bengio2003neural}
Bengio, Y., Ducharme, R., Vincent, P., Janvin, C.: A neural probabilistic
  language model. The journal of machine learning research  \textbf{3},
  1137--1155 (2003)

\bibitem{bollegala2010bottom}
Bollegala, D., Okazaki, N., Ishizuka, M.: A bottom-up approach to sentence
  ordering for multi-document summarization. Information processing \&
  management  \textbf{46}(1),  89--109 (2010)

\bibitem{bosselut-etal-2018-discourse}
Bosselut, A., Celikyilmaz, A., He, X., Gao, J., Huang, P.S., Choi, Y.:
  Discourse-aware neural rewards for coherent text generation. In: Proceedings
  of the 2018 Conference of the North {A}merican Chapter of the Association for
  Computational Linguistics: Human Language Technologies, Volume 1 (Long
  Papers). pp. 173--184. Association for Computational Linguistics, New
  Orleans, Louisiana (Jun 2018). \doi{10.18653/v1/N18-1016},
  \url{https://www.aclweb.org/anthology/N18-1016}

\bibitem{cer-etal-2018-universal}
Cer, D., Yang, Y., Kong, S.y., Hua, N., Limtiaco, N., St.~John, R., Constant,
  N., Guajardo-Cespedes, M., Yuan, S., Tar, C., Strope, B., Kurzweil, R.:
  Universal sentence encoder for {E}nglish. In: Proceedings of the 2018
  Conference on Empirical Methods in Natural Language Processing: System
  Demonstrations. pp. 169--174. Association for Computational Linguistics,
  Brussels, Belgium (Nov 2018). \doi{10.18653/v1/D18-2029},
  \url{https://www.aclweb.org/anthology/D18-2029}

\bibitem{chambers2008unsupervised}
Chambers, N., Jurafsky, D.: Unsupervised learning of narrative event chains.
  In: Proceedings of ACL-08: HLT. pp. 789--797 (2008)

\bibitem{chambers2009unsupervised}
Chambers, N., Jurafsky, D.: Unsupervised learning of narrative schemas and
  their participants. In: Proceedings of the Joint Conference of the 47th
  Annual Meeting of the ACL and the 4th International Joint Conference on
  Natural Language Processing of the AFNLP. pp. 602--610 (2009)

\bibitem{dehghani2018universal}
Dehghani, M., Gouws, S., Vinyals, O., Uszkoreit, J., Kaiser, {\L}.: Universal
  transformers. arXiv preprint arXiv:1807.03819  (2018)

\bibitem{devlin-etal-2019-bert}
Devlin, J., Chang, M.W., Lee, K., Toutanova, K.: {BERT}: Pre-training of deep
  bidirectional transformers for language understanding. In: Proceedings of the
  2019 Conference of the North {A}merican Chapter of the Association for
  Computational Linguistics: Human Language Technologies, Volume 1 (Long and
  Short Papers). pp. 4171--4186. Association for Computational Linguistics,
  Minneapolis, Minnesota (Jun 2019). \doi{10.18653/v1/N19-1423},
  \url{https://www.aclweb.org/anthology/N19-1423}

\bibitem{elsner2007unified}
Elsner, M., Austerweil, J., Charniak, E.: A unified local and global model for
  discourse coherence. In: Human Language Technologies 2007: The Conference of
  the North American Chapter of the Association for Computational Linguistics;
  Proceedings of the Main Conference. pp. 436--443 (2007)

\bibitem{gong2016end}
Gong, J., Chen, X., Qiu, X., Huang, X.: End-to-end neural sentence ordering
  using pointer network. arXiv preprint arXiv:1611.04953  (2016)

\bibitem{haddad2018handling}
Haddad, H., Fadaei, H., Faili, H.: Handling oov words in nmt using unsupervised
  bilingual embedding. In: 2018 9th International Symposium on
  Telecommunications (IST). pp. 569--574. IEEE (2018)

\bibitem{hochreiter1997long}
Hochreiter, S., Schmidhuber, J.: Long short-term memory. Neural computation
  \textbf{9}(8),  1735--1780 (1997)

\bibitem{jozefowicz2016exploring}
Jozefowicz, R., Vinyals, O., Schuster, M., Shazeer, N., Wu, Y.: Exploring the
  limits of language modeling. arXiv preprint arXiv:1602.02410  (2016)

\bibitem{kiros2015skip}
Kiros, R., Zhu, Y., Salakhutdinov, R.R., Zemel, R., Urtasun, R., Torralba, A.,
  Fidler, S.: Skip-thought vectors. In: Advances in neural information
  processing systems. pp. 3294--3302 (2015)

\bibitem{kozielski2013open}
Kozielski, M., Rybach, D., Hahn, S., Schl{\"u}ter, R., Ney, H.: Open vocabulary
  handwriting recognition using combined word-level and character-level
  language models. In: 2013 IEEE International Conference on Acoustics, Speech
  and Signal Processing. pp. 8257--8261. IEEE (2013)

\bibitem{kumar2020deep}
Kumar, P., Brahma, D., Karnick, H., Rai, P.: Deep attentive ranking networks
  for learning to order sentences. In: AAAI. pp. 8115--8122 (2020)

\bibitem{lapata2003probabilistic}
Lapata, M.: Probabilistic text structuring: Experiments with sentence ordering.
  In: Proceedings of the 41st Annual Meeting of the Association for
  Computational Linguistics. pp. 545--552 (2003)

\bibitem{lapata2006automatic}
Lapata, M.: Automatic evaluation of information ordering: Kendall's tau.
  Computational Linguistics  \textbf{32}(4),  471--484 (2006)

\bibitem{li-jurafsky-2017-neural}
Li, J., Jurafsky, D.: Neural net models of open-domain discourse coherence. In:
  Proceedings of the 2017 Conference on Empirical Methods in Natural Language
  Processing. pp. 198--209. Association for Computational Linguistics,
  Copenhagen, Denmark (Sep 2017). \doi{10.18653/v1/D17-1019},
  \url{https://www.aclweb.org/anthology/D17-1019}

\bibitem{lin2004automatic}
Lin, C.Y., Och, F.J.: Automatic evaluation of machine translation quality using
  longest common subsequence and skip-bigram statistics. In: Proceedings of the
  42nd Annual Meeting of the Association for Computational Linguistics
  (ACL-04). pp. 605--612 (2004)

\bibitem{logeswaran2018sentence}
Logeswaran, L., Lee, H., Radev, D.: Sentence ordering and coherence modeling
  using recurrent neural networks. In: Proceedings of the AAAI Conference on
  Artificial Intelligence. vol.~32 (2018)

\bibitem{medsker2001recurrent}
Medsker, L.R., Jain, L.: Recurrent neural networks. Design and Applications
  \textbf{5} (2001)

\bibitem{journals/corr/abs-1301-3781}
Mikolov, T., Chen, K., Corrado, G., Dean, J.: Efficient estimation of word
  representations in vector space. CoRR  \textbf{abs/1301.3781} (2013),
  \url{http://dblp.uni-trier.de/db/journals/corr/corr1301.html#abs-1301-3781}

\bibitem{mikolov2011empirical}
Mikolov, T., Deoras, A., Kombrink, S., Burget, L., {\v{C}}ernock{\`y}, J.:
  Empirical evaluation and combination of advanced language modeling
  techniques. In: Twelfth annual conference of the international speech
  communication association (2011)

\bibitem{mostafazadeh2017event}
Mostafazadeh, N.: From Event to Story Understanding. University of Rochester
  (2017)

\bibitem{mostafazadeh2016corpus}
Mostafazadeh, N., Chambers, N., He, X., Parikh, D., Batra, D., Vanderwende, L.,
  Kohli, P., Allen, J.: A corpus and cloze evaluation for deeper understanding
  of commonsense stories. In: Proceedings of the 2016 Conference of the North
  American Chapter of the Association for Computational Linguistics: Human
  Language Technologies. pp. 839--849 (2016)

\bibitem{mostafazadeh2016caters}
Mostafazadeh, N., Grealish, A., Chambers, N., Allen, J., Vanderwende, L.:
  Caters: Causal and temporal relation scheme for semantic annotation of event
  structures. In: Proceedings of the Fourth Workshop on Events. pp. 51--61
  (2016)

\bibitem{mostafazadeh2017lsdsem}
Mostafazadeh, N., Roth, M., Louis, A., Chambers, N., Allen, J.: Lsdsem 2017
  shared task: The story cloze test. In: Proceedings of the 2nd Workshop on
  Linking Models of Lexical, Sentential and Discourse-level Semantics. pp.
  46--51 (2017)

\bibitem{parikh-etal-2016-decomposable}
Parikh, A., T{\"a}ckstr{\"o}m, O., Das, D., Uszkoreit, J.: A decomposable
  attention model for natural language inference. In: Proceedings of the 2016
  Conference on Empirical Methods in Natural Language Processing. pp.
  2249--2255. Association for Computational Linguistics, Austin, Texas (Nov
  2016). \doi{10.18653/v1/D16-1244},
  \url{https://www.aclweb.org/anthology/D16-1244}

\bibitem{pichotta-mooney-2016-using}
Pichotta, K., Mooney, R.J.: Using sentence-level {LSTM} language models for
  script inference. In: Proceedings of the 54th Annual Meeting of the
  Association for Computational Linguistics (Volume 1: Long Papers). pp.
  279--289. Association for Computational Linguistics, Berlin, Germany (Aug
  2016). \doi{10.18653/v1/P16-1027},
  \url{https://www.aclweb.org/anthology/P16-1027}

\bibitem{pour2020new}
Pour, M.G., Razavi, S.Z., Faili, H.: A new sentence ordering method using bert
  pretrained model. In: 2020 11th International Conference on Information and
  Knowledge Technology (IKT). pp. 132--138. IEEE (2020)

\bibitem{schuster1997bidirectional}
Schuster, M., Paliwal, K.K.: Bidirectional recurrent neural networks. IEEE
  transactions on Signal Processing  \textbf{45}(11),  2673--2681 (1997)

\bibitem{sennrich-etal-2016-neural}
Sennrich, R., Haddow, B., Birch, A.: Neural machine translation of rare words
  with subword units. In: Proceedings of the 54th Annual Meeting of the
  Association for Computational Linguistics (Volume 1: Long Papers). pp.
  1715--1725. Association for Computational Linguistics, Berlin, Germany (Aug
  2016). \doi{10.18653/v1/P16-1162},
  \url{https://www.aclweb.org/anthology/P16-1162}

\bibitem{10.5555/2969033.2969173}
Sutskever, I., Vinyals, O., Le, Q.V.: Sequence to sequence learning with neural
  networks. In: Proceedings of the 27th International Conference on Neural
  Information Processing Systems - Volume 2. p. 3104–3112. NIPS'14, MIT
  Press, Cambridge, MA, USA (2014)

\bibitem{Wang2020SBERTWKAS}
Wang, B., Kuo, C.C.J.: Sbert-wk: A sentence embedding method by dissecting
  bert-based word models. IEEE/ACM Transactions on Audio, Speech, and Language
  Processing  \textbf{28},  2146--2157 (2020)

\bibitem{wang2019hierarchical}
Wang, T., Wan, X.: Hierarchical attention networks for sentence ordering. In:
  Proceedings of the AAAI Conference on Artificial Intelligence. vol.~33, pp.
  7184--7191 (2019)

\bibitem{woodland2002large}
Woodland, P.C., Povey, D.: Large scale discriminative training of hidden markov
  models for speech recognition. Computer Speech \& Language  \textbf{16}(1),
  25--47 (2002)

\end{thebibliography}

\end{document}